# Parole de présidents (1958-2022)

Dominique Labbé[1], Jacques Savoy[2]

[1] Université Grenoble Alpes - Dominique.Labbe@umrpacte.fr
[2] Université de Neuchâtel - Jacques.Savoy@unine.ch

**Abstract**

Over the past sixty-six years, eight presidents successively headed the Fifth French Republic (de Gaulle, Pompidou, Giscard d'Estaing, Mitterrand, Chirac, Sarkozy, Holland, Macron). After presenting the corpus of their speeches - 9,202 texts and more than 20 million labelled words - the style of each of them will be characterized by their vocabulary (lemmas and part-of-speech). A deeper analysis reveals the typical sequences of each tenant of the Elysée. Based on an intertextual distance between all presidential speeches, a synthesis can be drawn reflecting the similarities and differences between presidents.

**Résumé**

En plus de soixante ans, huit présidents se sont succédé à la tête de la V$^e$ République française (de Gaulle, Pompidou, Giscard d'Estaing, Mitterrand, Chirac, Sarkozy, Hollande, Macron). Après avoir présenté le corpus de leurs discours - soit 9 202 textes et plus de 20 millions de mots étiquetés - le style de chacun des présidents sera caractérisé à l'aide de leurs vocabulaire (vocables et catégories grammaticales). Une analyse plus approfondie révèle les séquences typiques de chaque locataire de l'Élysée. Basée sur les distances entre l'ensemble des allocutions, une figure illustre les similitudes et différences entre les différents présidents.

**Mots clés :** Stylométrie – Présidents français – Vocabulaire caractéristique – Partie du discours – Séquences de mots – Distance intertextuelle entre auteurs.

## 1. Introduction

Depuis l'avènement de la V$^e$ République en 1958, le président exerce un rôle de premier plan dans la vie politique française (Minc 2020), (Perrineau 2024). Cette fonction possède de fortes similarités avec celle des Etats-Unis. Dans les deux cas, le président est le chef de l'État, le haut responsable de l'administration, le chef des armées, le leader d'un parti tout en dirigeant l'exécutif (fonction partagée avec le premier ministre). On soulignera que le président garde la haute main sur la politique étrangère, politique possédant depuis le général de Gaulle une certaine indépendance comparée à d'autres pays européens. De plus, le président est le seul homme politique élu par l'ensemble des citoyens du pays. La fonction présidentielle française se situe donc au cœur du pouvoir, ce qui la distingue de celle qui prévaut dans un système parlementaire comme en Allemagne ou en Italie. Enfin on doit garder à l'esprit que le rôle de l'État français s'avère plus important que dans d'autres pays européens.

De 1958 à 2024, huit présidents se sont succédés au pouvoir, à savoir de Gaulle (1958–1969), Pompidou (1969–1974), Giscard d'Estaing (1974–1981), Mitterrand (1981–1995), Chirac (1995–2007), Sarkozy (2007–2012), Hollande (2012–2017) et Macron (2017–). Bien que la politique impose un certain jargon, chaque locuteur se caractérise par des mots et un style qui lui sont propres. Et pour un chef de l'exécutif, parler au peuple, aux médias ou devant une assemblée s'avère un exercice presque quotidien comme le souligne Caesar *et al.* (1981) qui affirme que "*speaking is governing*". On peut certes ajouter que certains présidents parlent davantage que d'autres (Minc 2020).

Grâce à l'informatique, on dispose d'études sur le vocabulaire d'un auteur ou d'un groupe d'auteurs (Savoy 2020). Le discours politique a également été étudié pour lui-même. En se centrant sur le discours du gouvernement en français, Labbé & Monière (2003; 2008) ont créé un corpus comprenant les *Discours du Trône* (Canada et Québec), ceux de politique générale (France) ou la communication des principaux candidats lors de l'élection présidentielle 2012 (Labbé & Monière 2013), (Arnold & Labbé 2015), (Arnold *et al*. 2016). En se focalisant sur un leader, des études ont été menées sur quelques présidents comme de Gaulle (Cotterret & Moreau 1969), (Arnold 2019), Mitterrand (Labbé 1990), Chirac (Mayaffre 2004) ou Sarkozy (Mayaffre 2012). Des recherches similaires ont été entreprises concernant les présidents italiens (Pauli & Tuzzi 2009), tchèques (Kubát & Cech 2016) ou américains (Hart 1984), (Hart *et al*. 2013), (Savoy 2015).

De ces analyses, on peut conclure que les institutions tendent à effacer partiellement les différences entre partis lorsque leurs leaders exercent le pouvoir. Le facteur temps tend aussi à expliquer les variations entre présidents ou premiers ministres. La prise de pouvoir d'un leader très charismatique peut s'accompagner d'un changement notable de vocabulaire (Labbé & Monière 2003) de même que des évènements exceptionnels (guerre mondiale, dépression économique) (Savoy, 2017).

La suite de cet article prolonge nos précédentes publications (Labbé *et al*. 2020), (Labbé & Savoy 2021) et s'organise de la façon suivante. La deuxième section expose les grandes lignes du corpus généré pour nos analyses. La troisième section présente le vocabulaire de chaque président et les vocables qui les caractérisent. La quatrième se focalise sur les séquences de mots caractérisant le style de chaque président. La cinquième dresse une carte des similarités entre les huit présidents. Une conclusion propose une synthèse de nos analyses.

## 2. Corpus des discours présidentiels

Les discours de de Gaulle et de Pompidou ont été rassemblés par D. Labbé à partir des archives et d'extraits publiés en livre. A partir de Giscard d'Estaing (1974), des archives électroniques ont été constituées par la présidence. Le texte des discours présidentiels a été téléversé du site web de l'Elysée (www.elysee.fr) ou du site *Vie Publique* (www.vie-publique.fr). Pour chaque allocution, un prétraitement a été effectué afin d'obtenir la même orthographe pour des variations possibles (par exemple, Abu Dabi, Abou Dhabi, Abou-Dabi, etc.). Ensuite, suivant un processus semi-automatique, une étiquette indiquant la catégorie grammaticale (e.g., adjectif, pronom personnel 1$^{\text{ère}}$ personne singulier) et le lemme (entrée dans le dictionnaire) a été attribuée à chaque mot. Ces deux informations forment un vocable. On doit se rappeler que la grammaire française dispose de nombreuses variations verbales (Sproat 1992). Dans la suite de cet article, nos statistiques sont basées sur les vocables.

En analysant un document rédigé en français, on doit se souvenir que cette langue connaît de nombreuses ambiguïtés rendant tout processus difficile à automatiser complètement. Par exemple, dans la phrase « Je suis le président », le verbe 'suis' peut indiquer le verbe 'être' (le sens habituel) mais également le verbe 'suivre' (si la phrase est prononcée par un ministre suivant le président). Autre exemple avec la phrase « On vit des évènements graves » dans laquelle le verbe 'vit' peut signifier 'voir' (au passé simple) ou 'vivre' (indicatif présent).

Enfin on gardera à l'esprit que les interventions publiques présidentielles se sont déroulées dans des cadres très divers. Dès lors, outre la question du genre (allocutions, conférences de presse, discours électoraux, etc.), certaines caractéristiques de ces textes peuvent refléter l'influence des circonstances dans lesquels ils ont été émis. Pour faire apparaître les caractéristiques propres au style de chaque président, il faudrait pouvoir aussi neutraliser ces facteurs exogènes en étudiant des interventions présentées dans un cadre à peu près semblable. Par exemple, les discours sur *State of Union* que prononcent les présidents américains chaque année depuis la fondation des Etats-Unis (Savoy 2015; 2017) ou les messages de vœux des présidents français (Leblanc 2016), (Labbé *et al*. 2020).

Quelques statistiques décrivant les particularités essentielles de notre corpus sont présentées dans le tableau 1. Sous l'étiquette 'Président', nous avons signalé les années de la présidence puis le nombre total de textes (Discours), le nombre moyen par année (Par an) et la longueur moyenne d'une allocution par chaque président (Moyenne). Enfin les deux colonnes suivantes indiquent le nombre total de mots (Longueur) et la taille du vocabulaire (nombre de vocables différents).

Une dernière ligne révèle que ce corpus comprend 9 202 textes pour un total de plus de 20 millions de mots. Les trois tomes des *Mémoire de guerre* du général de Gaulle comptent 372 644 mots. Si l'on imprimait toutes les interventions des présidents au même format, il faudrait 161 volumes, soit plus de trois mètres linéaires de rayonnage.

Tableau 1. Principales caractéristiques du corpus des discours présidentiels

|  | Président | Discours | Par an | Moyenne | Longueur | Vocabulaire |
|---|---|---|---|---|---|---|
| De Gaulle* | 1958-1969 | 460 | 44 | 908 | 417 146 | 8 663 |
| Pompidou* | 1969-1974 | 137 | 28 | 1 923 | 263 309 | 7 736 |
| Giscard d'Estaing* | 1974-1981 | 191 | 27 | 3 504 | 669 332 | 9 062 |
| Mitterrand | 1981-1995 | 2 547 | 182 | 2 217 | 5 643 845 | 23 834 |
| Chirac | 1995-2007 | 2 478 | 207 | 1 668 | 4 131 849 | 22 558 |
| Sarkozy | 2007-2012 | 1 074 | 215 | 3 035 | 3 256 896 | 20 049 |
| Hollande | 2012-2017 | 1 545 | 309 | 2 093 | 3 231 971 | 19 146 |
| Macron | 2017-2022 | 770 | 110 | 3 135 | 2 412 712 | 20 615 |
| Total | 1958-2022 | 9 202 |  | 2 178 | 20 027 931 | 44 072 |

\* Corpus incomplets (en cours de constitution)

Avant 1981, les interventions des présidents n'ont pas été systématiquement archivées. Les corpus du Général, de Pompidou et de Giscard d'Estaing ne sont pas encore complets. Nous disposons également d'un nombre plus élevé d'allocutions de Mitterrand et Chirac car leurs mandats à l'Élysée – respectivement quatorze et douze ans – ont été plus longs que ceux des autres présidents.

Avec les années, les présidents tendent à prononcer des allocutions plus longues, culminant avec Macron (moyenne de 3 135 mots / discours). Les deux premiers chefs d'État parlaient moins longtemps (e.g., de Gaulle, moyenne de 908) suggérant une plus grande concision. Toutefois, une étude précédente (Labbé & Savoy 2021) démontrait que leurs discours présentaient aussi une richesse lexicale plus élevée impliquant des allocutions pas forcément plus aisées à comprendre.

Enfin, il convient de signaler que la fonction présidentielle s'est modifiée durant cette période. Dès les années 1980 le chef de l'exécutif communique davantage. Durant la même période, Reagan a été surnommé le « *Great Communicator* » aux Etats-Unis (Hart 1984). Ce phénomène se propage aussi en France avec Mitterrand prononçant, en moyenne, 182 allocutions chaque année. Cette tendance se renforce sous Chirac (207 discours / an) puis Hollande (309). Ce phénomène illustre d'abord la présidentialisation croissante des institutions françaises. Elle peut aussi s'expliquer par l'importance accrue d'Internet, des réseaux sociaux, la multiplication des chaînes d'information et des sondages mais aussi la volonté des leaders politiques de s'adresser directement au peuple. Toutefois la personnalité joue aussi un rôle. Par exemple, Macron préfère réduire cette visibilité de même que le nombre de conférences de presse (en limitant aussi le nombre de questions auxquelles il répond).

## 3. Le vocabulaire caractéristique de chaque président

Afin de décrire le style propre à chaque auteur, Biber & Conrad (2009) suggèrent de tenir compte du vocabulaire usité et, en particulier, des mots les plus fréquents. Cependant, extraire les formes les plus employées (de, la, et, l, à, le, les, que, est, des, d, …) ou les vocables (le, de, être, à, et, que, ce, avoir, un, je, …) n'aboutit pas à contraster le style de chaque auteur. Souvent cette information ne confirme que la présence de la langue française et du vocabulaire politique (e.g., 'France' apparait en 31e rang, 'pays' en 36e). Par comparaison, dans un corpus d'articles de presse, les mots 'police', 'groupe', 'millions' ou 'années' possèdent des fréquences plus élevées que dans les discours présidentiels.

Si tous les présidents recourent généralement aux mêmes mots, la différence stylistique entre eux réside dans les fréquences d'occurrence. Ainsi certains leaders usent plus abondamment de certains vocables et tendent à négliger d'autres. Afin de déterminer les vocables suremployés par un locuteur, Muller (1977) suggère de comparer les fréquences d'apparition chez un auteur comparé aux autres. Par exemple, le lemme 'France' apparait 6 606 fois chez Macron (soit une fréquence d'occurrence de 2,3‰) mais 2 480 fois chez de Gaulle (5,3‰) soit plus du double. Dès lors peut-on affirmer que le vocable est suremployé par le Général et sous-employé par Macron ? Une description de la méthode utilisée pour répondre à cette question est décrite dans (Labbé & Savoy 2021). Les principaux lemmes suremployés par chaque président sont présentés dans le tableau 2.

Tableau 2. Quelques vocables caractéristiques de chaque président

| Président | 1er vocable | 2e vocable | 3e vocable | 4e vocable |
|---|---|---|---|---|
| De Gaulle | honneur | Algérie | univers | peuple |
| Pompidou | Nixon | conséquent | parisien | dévaluation |
| Giscard d'Estaing | actuel | soixante | détente | hausse |
| Mitterrand | je | bien | neuf | entendu |
| Chirac | naturellement | le | mondialisation | européen |
| Sarkozy | on | ne pas | crise | G20 |
| Hollande | pouvoir | nous | être / avoir | ici |
| Macron | nous / notre | sujet | transformation | engagement |

Ces termes révèlent les thèmes récurrents d'une présidence (Algérie, dévaluation, détente, mondialisation) et des aspects stylistiques (je, nous, bien, ne pas, ici). Pour Sarkozy, président durant la crise économique de 2008, la solution à cette crise semble être du côté du G20. On notera son usage récurrent du pronom impersonnel 'on' et de la négation. Son successeur favorise le terme 'pouvoir' et les verbes auxiliaires être et avoir. Après Mitterrand et le pronom 'je/moi' ou le 'on' de Sarkozy, Hollande met en avant le 'nous' qui lui permet d'établir un lien avec l'auditeur mais aussi de fondre le locuteur dans le groupe national qui porte la responsabilité des difficultés. Macron reste dans cette ligne mais en renforçant l'usage du 'nous' (et 'notre'). De plus, chez ce président des réformes, les mots 'transformation' et 'engagement' signalent son souci de changer la société, et indiquent ses engagements 'd'hyper-président'.

Pour distinguer les divers styles, on peut également recourir aux fréquences d'usage des verbes modaux. Ces verbes (devoir, pouvoir, vouloir, falloir) sont essentiels pour exprimer divers degrés de certitude, de permission, de souhait ou d'obligation. A cette liste, nous avons ajouté le verbe 'faire'.

Comme l'indique le tableau 3, chaque occupant de l'Élysée tend à favoriser un verbe plutôt qu'un autre. Comparé aux autres, de Gaulle recourt davantage aux verbes 'vouloir' et 'devoir'. Pour Sarkozy et Hollande, le 'falloir' est une forme modale privilégiée car elle est impersonnelle (que confirment la préférence pour le 'on' de Sarkozy et pour le 'nous' chez Hollande). Avec les années, on constate une augmentation de l'emploi du verbe 'pouvoir' qui s'avère le plus important chez Hollande et Macron. Si Hollande demeure le leader dans l'emploi du verbe 'faire', Sarkozy et Macron ont également tendance à en user de manière récurrente, ce qui n'était pas le cas de leurs prédécesseurs. En résumé, le tableau 3 signale que les trois derniers présidents sont des adeptes de l'usage du verbe 'pouvoir' ainsi que 'faire'. Cette caractéristique stylistique s'affirme avec les années.

Tableau 3. Emploi des verbes modaux par chaque président (en pour mille mots)

| Président | pouvoir | falloir | vouloir | devoir | faire |
|---|---|---|---|---|---|
| De Gaulle | 0,08‰ | 0,00‰ | 0,13‰ | 0,07‰ | 1,36‰ |
| Pompidou | 0,20‰ | 0,00‰ | 0,06‰ | 0,02‰ | 1,44‰ |
| Giscard d'Estaing | 0,19‰ | 0,01‰ | 0,04‰ | 0,01‰ | 1,78‰ |
| Mitterrand | 0,25‰ | 0,01‰ | 0,08‰ | 0,03‰ | 1,51‰ |
| Chirac | 0,22‰ | 0,01‰ | 0,06‰ | 0,02‰ | 1,74‰ |
| Sarkozy | 0,24‰ | 0,03‰ | 0,08‰ | 0,02‰ | 2,21‰ |
| Hollande | 0,46‰ | 0,02‰ | 0,05‰ | 0,01‰ | 2,82‰ |
| Macron | 0,48‰ | 0,01‰ | 0,05‰ | 0,03‰ | 2,10‰ |

## 4. Parties du discours et séquences caractéristiques de chaque président

Dans la section précédente, nous avons défini une méthode afin de déterminer les vocables caractéristiques de chaque président (Labbé & Savoy, 2021). Une approche identique s'applique afin de définir les catégories grammaticales (voir le tableau 4) ou brèves séquences de vocables (tableau 5). Même si le corpus possède un volume conséquent, les séquences répétées de quatre ou cinq termes connaissent des fréquences d'occurrence faibles. Établir une analyse sur de telles représentations s'avèrent par trop hasardeux.

Afin de distinguer les styles de chaque président, le tableau 4 reprend les parties du discours (*part-of-speech*, POS) mais en tenant compte de certains traits morphologiques, en particulier pour les verbes et pronoms. De Gaulle recourt aux verbes au passé simple et, comparé aux autres présidents, préfère les déterminants possessifs (e.g., notre, mon). Ses allocutions contiennent plus de ponctuation, c'est-à-dire des virgules et points-virgules. Ce dernier signe de ponctuation disparait quasiment des discours ultérieurs. Pompidou favorise les adjectifs et les conjonctions construisant des phrases plus complexes à comprendre pour l'auditeur. Le changement de style sera plus marqué avec Giscard d'Estaing qui favorise les nombres et locutions. Notons également que le temps des verbes demeure tourné vers le passé (indicatif imparfait).

Pour Mitterrand, le pronom personnel (e.g., je, moi, vous) tient une place centrale de même que les adverbes (e.g., bien). Son successeur se tourne vers le groupe nominal (article, adjectif, nom) adoptant ainsi un style plus descriptif et explicatif. Pour Sarkozy, le discours doit être orienté vers l'action (verbe) à l'indicatif présent. On note également que ce président ancre son intervention avec un usage plus important de noms de personnes ou de lieu. Enfin, il cherche à impliquer plus l'auditeur par le recours à des pronoms personnels (nous, je, tu).

Si Hollande favorise les participes passés, c'est qu'il parle souvent à la voix passive mais aussi parce que le passé composé est le temps qu'il préfère. Il recourt également plus aux verbes à l'infinitif qui entrent dans les constructions modales (e.g., 'il faut faire'). Enfin Macron rédige de longues phrases dont la construction requiert des conjonctions de coordination (e.g., et, ou, mais). On peut également signaler la profusion de déterminants démonstratifs (e.g., ce, cette) caractéristiques de discours à visée pédagogique, et de mots d'origine étrangère, en particulier en anglais (e.g., *too big to fail, power, smart, fund*).

Tableau 4. Les trois catégories grammaticales les plus caractéristiques de chaque président

| Président | 1er | 2e | 3e |
|---|---|---|---|
| De Gaulle | Verbe, passé simple | Déterminant possessif | Ponctuation |
| Pompidou | Conj. coord. / subor. | Adjectif | Adverbe |
| Giscard d'Estaing | Nombre | Locution | Verbe, imparfait |
| Mitterrand | Pronom personnel | Adverbe | Conj. subordination |
| Chirac | Adjectif | Nom commun | Article |
| Sarkozy | Verbe, présent | Nom propre | Pronom personnel |
| Hollande | Verbe, infinitif | Verbe, participe passé | Autres pronoms |
| Macron | Déter. démonstratif | Conj. coordination | Mots étrangers |

Afin de cerner au plus près les traits caractéristiques du vocabulaire de chaque locuteur, on peut tenir compte non pas seulement des termes (vocables ou formes) mais des séquences brèves composées de deux ou trois mots. Ainsi le vocable isolé 'union' ne désigne pas clairement l'intention de l'auteur. Parle-t-on de l'Union Européenne, l'Union soviétique ou d'autres formes (union sacrée, économique, …). Autre exemple avec 'hausse' dans le tableau 2 associé avec Giscard d'Estaing, une hausse de quoi ? Des prix (la bonne interprétation), du coût du pétrole, des impôts ? Une séquence un peu plus longue lèverait cette ambiguïté.

Dans ce but, nous avons repris notre corpus et détecté les bigrammes ou trigrammes les plus caractéristiques de chaque président. De plus certaines séquences ne présentent presque aucun intérêt comme celles comprenant " de le ", " l' ", ou " d' ". Elles peuvent être ignorées.

Tableau 5. Séquence de deux ou trois mots parmi les plus caractéristiques de chaque président

| Président | n-gramme | n-gramme | n-gramme |
|---|---|---|---|
| De Gaulle | en l'honneur de | le monde libre | autrement dit |
| Pompidou | moins vrai que | le marché commun | parité fixe |
| Giscard d'Estaing | à l'heure actuelle | de la détente | la hausse de |
| Mitterrand | tiers-monde | la communauté européenne | bien entendu |
| Chirac | l'Union Européenne | ce qui concerne | la mondialisation |
| Sarkozy | le G 20 | dire une chose | la crise |
| Hollande | faire en sorte | que nous pouvons | pouvoir être |
| Macron | sur ce sujet | je veux ici | en la matière |

Le tableau 5 reprend quelques-uns des bi- ou trigrammes les plus caractéristiques de chaque chef d'État. De Gaulle lève son verre en l'honneur de son invité(e), et compare les deux blocs de l'époque avec le monde libre et celui sous domination de la Russie. Le tableau 5 signale également des tournures de phrases caractérisant chaque locataire de l'Élysée. Le Général emploie le 'autrement dit', Pompidou opte pour '(pas) moins vrai que', Giscard choisit 'à l'heure actuelle' et Mitterrand le 'bien entendu'. Pour les plus récents, on reconnaitra le 'ce qui concerne', 'dire une chose', 'faire en sorte' ou 'je veux ici'.

Mais la France est tournée vers l'Europe dont la dénomination varie d'une présidence à l'autre. Ainsi, Pompidou parlait de 'marché commun' tandis que Mitterrand de 'communauté européenne', élargissant l'Europe de son premier objet, l'économie, vers d'autres thèmes. Chirac a préféré la dénomination 'Union Européenne'. Enfin, si Hollande suremployait le lemme 'pouvoir' (voit tableau 2), le tableau 5 signale que cet usage est associé au verbe, et non au substantif.

Pour déterminer des séquences répétitives associées à chaque président, nous pouvons également extraire les suites de deux vocables les plus fréquents correspondant à un patron syntaxique. Dans ce cas, seulement les noms, adjectifs et verbes sont retenus comme segments d'extraction. Toutefois, quelques séquences reviennent presque pour tous les présidents comme 'Monsieur (le) président', 'président (de la) République', ' Madame, Monsieur' ou 'France être' (pour 'la France est…').

Tableau 6. Séquence de deux vocables parmi les plus fréquent de chaque président

| Président | 1er | 2e | 3e | 4e |
|---|---|---|---|---|
| De Gaulle | être vrai | peuple français | pouvoir être | chef État |
| Pompidou | vouloir dire | général Gaulle | être vrai | devoir être |
| Giscard d'Estaing | vouloir dire | être fait | être important | devoir être |
| Mitterrand | être vrai | vouloir dire | avoir besoin | être bon |
| Chirac | être vrai | devoir être | avoir besoin | être fait |
| Sarkozy | avoir besoin | vouloir dire | devoir être | avoir droit |
| Hollande | pouvoir être | faire sorte | devoir être | être capable |
| Macron | avoir besoin | être train | devoir être | vouloir dire |

Le tableau 6 regroupe les quatre segments de deux vocables les plus récurrents sous chaque présidence. Peu de syntagmes nominaux reviennent dans ces segments fréquents comme 'général (de) Gaulle' ou 'chef (d')État'. Les verbes modaux et auxiliaires apparaissent fréquemment pour dénoter le degré de certitude ou la possibilité (e.g., 'être vrai', 'pouvoir être'). Le discours présidentiel doit être explicatif ou justifie une action ou une proposition de loi. Ainsi le verbe 'dire' revient dans les paroles de plusieurs présidents (de Pompidou à Macron) sous la forme de 'vouloir dire' permettant au locuteur de bien préciser sa pensée.

Les présidents doivent également signaler l'importance d'un objet ('avoir besoin') ou dénoter son statut ('être important', 'être bon'). Enfin, certaines séquences sont liées à un président particulier comme 'peuple français' pour le Général, 'général (de) Gaulle' pour Pompidou, 'avoir droit' pour Sarkozy, 'être capable' ou 'faire (en) sorte' pour Hollande ou 'être (en) train' pour Macron.

## 5. Distance intertextuelle entre présidents

Afin de dresser une image plus synthétique des similitudes et différences entre les allocutions des présidents, nous nous proposons de calculer une distance intertextuelle entre chaque paire de documents. Ces derniers regrouperont tous les discours attribués à un président. Toutefois, nous avons subdivisé les allocutions de Mitterrand et Chirac en deux, chacun correspondant à un mandat à l'Élysée.

Cette méthode est décrite de manière détaillée dans (Labbé & Labbé 2006). Dans les grandes lignes, la distance entre deux écrits se situe entre 0 (les deux n'ont rien en commun) à 1.0 (deux textes sont identiques). Cette valeur se calcule selon le taux de recouvrement ; si les deux documents possèdent de nombreux mots communs avec des fréquences d'occurrence proches, la distance sera faible. Si le vocabulaire s'avère très distinct et les fréquences d'apparition dissemblables, la distance retournée sera élevée.

En appliquant cette approche sur chaque paire de profils présidentiels, nous obtenons une matrice symétrique (10 x 10 = 100 valeurs) rendant difficile l'interprétation. Pour dégager les proximités, on applique à ces données une classification automatique (Baayen 2008) selon une technique dérivée des arbres phylogénétiques (Paradis 2011).

La figure 1 illustre le résultat obtenu nous permettant de mieux visualiser les relations entre présidents. Dans cette illustration, la distance correspond à la longueur des traits requis pour joindre une paire d'auteur. Par exemple, pour rejoindre Mitterrand 1 (1981-1988) à Mitterrand 2 (1988-1995), on constate que la longueur des deux traits demeure très faible, signalant un fort rapprochement stylistique. La même remarque s'applique pour la paire Chirac (1995-2002) et (2002-2007), ce qui permet de vérifier, au passage, la capacité de cet outil à reconnaître l'auteur d'un texte ou d'un groupe de textes. On remarque aussi que le style du général de Gaulle est le plus éloigné de tous les autres et que la distance la plus élevée se situe entre lui et Hollande.

Figure 1. Représentation arborée de la distance intertextuelle entre chaque président (1958-2022)

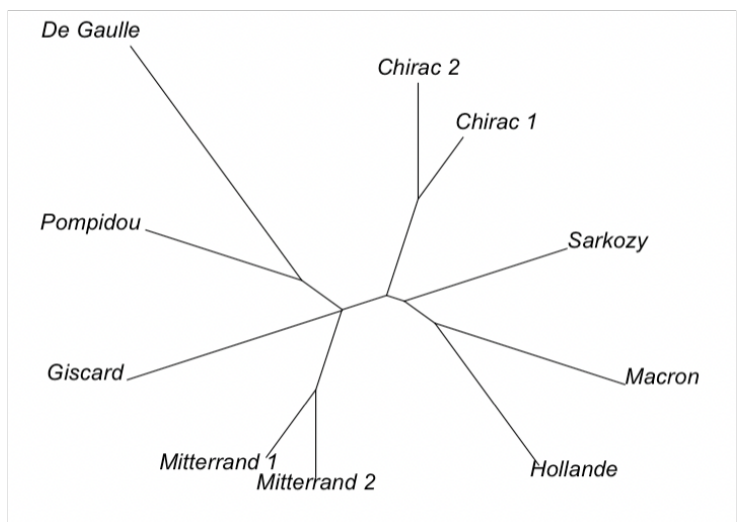

En général, cette figure illustre l'effet du temps. Plus les présidents sont séparés par les années, plus la distance grandit. Le style et les thèmes abordées par les allocutions présidentielles se différencient clairement avec le temps. De plus, chaque période ou décennie tend à favoriser un style plutôt qu'un autre. Ainsi, un leader politique ne s'adresse plus au peuple dans les mêmes termes ni avec le même style entre 1959 et 2022.

Dans la figure 1, on constate que Pompidou est celui qui se rapproche le plus du Général, puis un écart assez clair sépare cette paire de Giscard d'Estaing. Ce dernier appartient à un autre parti que ses deux prédécesseurs désirant s'éloigner du clan gaulliste. Il ouvre une autre politique requérant un style nouveau (Minc 2020), (Perrineau 2024). Le nouveau chef d'État doit marquer sa différence. Les deux chocs pétroliers (1973 et 1978) marquent la fin des trente glorieuses et l'apparition du chômage de masse.

En 1981, l'arrivée de Mitterrand au pouvoir marque une césure dans la V$^e$ République française. Premier socialiste à la tête de la V$^e$ République, il ouvre une trajectoire (hausse des salaires, nationalisations, élargissement de la couverture sociale) mais le coût économique le contraint vite à réduire ses ambitions sociales et à revenir à une politique de rigueur à partir de 1983. Cependant, le graphe indique que les deux mandats de Mitterrand s'inscrivent dans une relative continuité par rapport à ses trois prédécesseurs, notamment avec Giscard d'Estaing.

Le graphique indique que la coupure majeure survient en 1995 avec l'avènement au pouvoir de Chirac (1995-2007). Tous les discours antérieurs figurent à gauche de la figure et les suivants à droite. Ce deuxième volet de la V$^e$ République se caractérise par une intégration plus forte au niveau européen, la naissance de l'euro, et une influence plus marquée de la mondialisation. La présidence de Sarkozy (2007-2012) se rapproche de celle de Chirac, les deux provenant du même mouvement politique. Avec Hollande (2012-2017), on s'éloigne de ces présidents issus de la droite avec une politique suivant un modèle social-démocrate. Toutefois, sur l'arbre, le chemin qui relie Hollande à Mitterrand est trop long pour qu'on puisse envisager l'idée d'une "filiation" de l'un à l'autre. Enfin Macron, président qui se dit " ni de droite ni de gauche ", présente une affinité avec Hollande (du moins jusqu'en 2022).

## 6. Conclusion

Étudier et analyser les styles d'écriture de divers auteurs repose sur quelques principes essentiels. Certes, il s'avère ardu de décrire *per se* un style, mais, par comparaison, il est possible de détecter ce qui est propre à un auteur et commun à plusieurs

Dans cette perspective, nous avons repris huit présidents afin de découvrir leurs caractéristiques distinctives. Quelques statistiques signalent qu'avec les années, les occupants de l'Élysée parlent plus souvent et que leur intervention moyenne tend à s'allonger. Ce trend n'est pas parfaitement aligné car la personnalité du président joue aussi un rôle (e.g., Macron). Cet aspect s'explique aussi par la personnalisation croissante de la politique, spécialement de la présidence, l'importance grandissante des médias, des réseaux sociaux et des chaînes d'information.

Deuxièmement, notre analyse permet de détecter les termes (vocables, partie du discours, brèves séquences de mots) caractéristiques d'un président comparé aux autres. Avec les années, on constate que la ponctuation (virgule, point-virgule) tend à se réduire dans les allocutions, signe probable d'une simplification syntaxique des phrases, du moins avant Macron.

L'emploi des pronoms personnels ne demeure pas stable avec les premiers chefs d'État de la V$^e$ République, Mitterrand favorise le 'je/moi/me' tandis que Hollande puis Macron insisteront davantage sur le 'nous' dont la signification est bien différente (Pennebaker 2011). Dans cette perspective, Sarkozy opte pour le pronom indéfini 'on', introduisant une plus grande incertitude quant aux personnes désignées dans son discours.

Les séquences brèves permettent de distinguer les tournures et expressions préférées de chaque locuteur. Ainsi, Pompidou tend à employer le 'pas moins vrai que', Giscard d'Estaing le 'à l'heure actuelle' et Mitterrand le 'bien entendu'. Cette technique permet également de lever l'incertitude lorsque l'analyse se limite aux vocables isolés.

Les verbes peuvent être sous-employés au profit des noms communs comme, par exemple, chez Chirac. Chez le Général, ils se conjuguent plus couramment au passé simple, à l'indicatif imparfait chez Giscard d'Estaing, à l'indicatif présent chez Sarkozy tandis que Hollande opte volontiers pour la voie passive ou le passé composé. Les verbes modaux nous indiquent une tendance à l'accroissement de l'usage de 'pouvoir' au fil des décennies. De Gaulle se distingue par un emploi abondant de 'vouloir' et 'devoir' mais leurs fréquences diminuent avec le temps.

Troisièmement, le recours à la distance intertextuelle et à la classification automatique permet de visualiser les similitudes entre présidents, ce qui permet de découper les années 1958-2022 en deux groupes, soit les quatre premiers présidents (de Gaulle, Pompidou, Giscard d'Estaing, Mitterrand) d'une part et, d'autre part, les quatre derniers (Chirac, Sarkozy, Hollande, Macron).

Cette analyse peut être complétée par le recours à d'autres sources comme la syntaxe avec l'étude de brèves séquences de différentes catégories grammaticales. Enfin, beaucoup d'autres mesures sont possibles comme l'étude du sens que donne chacun de ces locuteurs aux vocables les plus fréquents de la politique – comme France, pays, Europe, monde - ou encore celle de la longueur et de la structure des phrases qui feront l'objet de publications ultérieures.

## Remerciements

## Références